\documentclass[letterpaper,10pt,conference]{ieeeconf}

\IEEEoverridecommandlockouts
\usepackage{graphicx}
\usepackage{amsmath,amssymb}
\usepackage{booktabs}
\usepackage{multirow}
\usepackage{cite}
\usepackage{url}
\usepackage{xcolor}
\usepackage{xspace}
\usepackage{tikz}
\usetikzlibrary{arrows.meta,positioning,fit}

\newcommand{\method}{SECD\xspace}

\usepackage[hidelinks]{hyperref}
\title{\LARGE \bf
Shared Execution-Clock Drifting Policy\\
for Dynamic Precision Manipulation
}

\author{Zhenchen Dong$^{1,*}$, Qingran Wu$^{2,*}$, Jinna Fu$^{3}$, Jiaming Wu$^{4}$\\Fulin Chen$^{5}$, Hongyu Yu$^{1,\dagger}$, Yide Liu$^{3,\dagger}$
\thanks{$^{*}$Equal contribution. $^{\dagger}$Corresponding authors: Hongyu Yu and Yide Liu.}
\thanks{$^{1}$The Hong Kong Polytechnic University.}
\thanks{$^{2}$University of California San Diego.}
\thanks{$^{3}$Zhejiang University.}
\thanks{$^{4}$Universit\'e Paris-Saclay.}
\thanks{$^{5}$Shanghai University of Engineering Science.}
\thanks{\raggedright Correspondence: \texttt{hongyu.yu@polyu.edu.hk}; \texttt{yide\_liu@zju.edu.cn}.}
}

\begin{document}

\maketitle
\thispagestyle{empty}
\pagestyle{empty}

\begin{abstract}
Manipulation under time constraints requires both accurate actions and an execution rhythm that matches the evolving scene. This becomes critical when a robot must intercept moving objects or complete a sequence of adjustments before a deadline. Although one-step policies reduce generation cost, their directly predicted action sequences leave temporal allocation implicit. We propose \textbf{Shared Execution-Clock Drifting (\method)}, which makes execution rhythm an explicit part of one-step action generation. Conditioned on an observation and a latent sample, the policy jointly predicts a progress-indexed action curve and a shared monotone clock that maps fixed control times to locations on the curve. Demonstration-derived alignment anchors this decomposition, which is trained jointly through drifting on the decoded actions. The resulting policy retains a fixed-rate control interface and requires one network evaluation. We evaluate \method across four real-robot tasks with inference on NVIDIA Thor. Across 300 trials, it achieves 77.00\% task-averaged success and outperforms the evaluated one-step baselines on every task, including 91\% success in cup retrieval from a 16\,m/min conveyor and 54\% in restoring and folding a crumpled shirt within 90\,s. A fixed-clock variant reaches 79\% on the same conveyor protocol. Complementary state-based RoboMimic experiments, including cross-seed ablations on Transport and Square, further support the joint design of the temporal representation and demonstration alignment. Project page: \url{https://secd-anonymous-ewn.pages.dev/}.
\end{abstract}

\begin{figure*}[t]
\centering
\includegraphics[width=\textwidth]{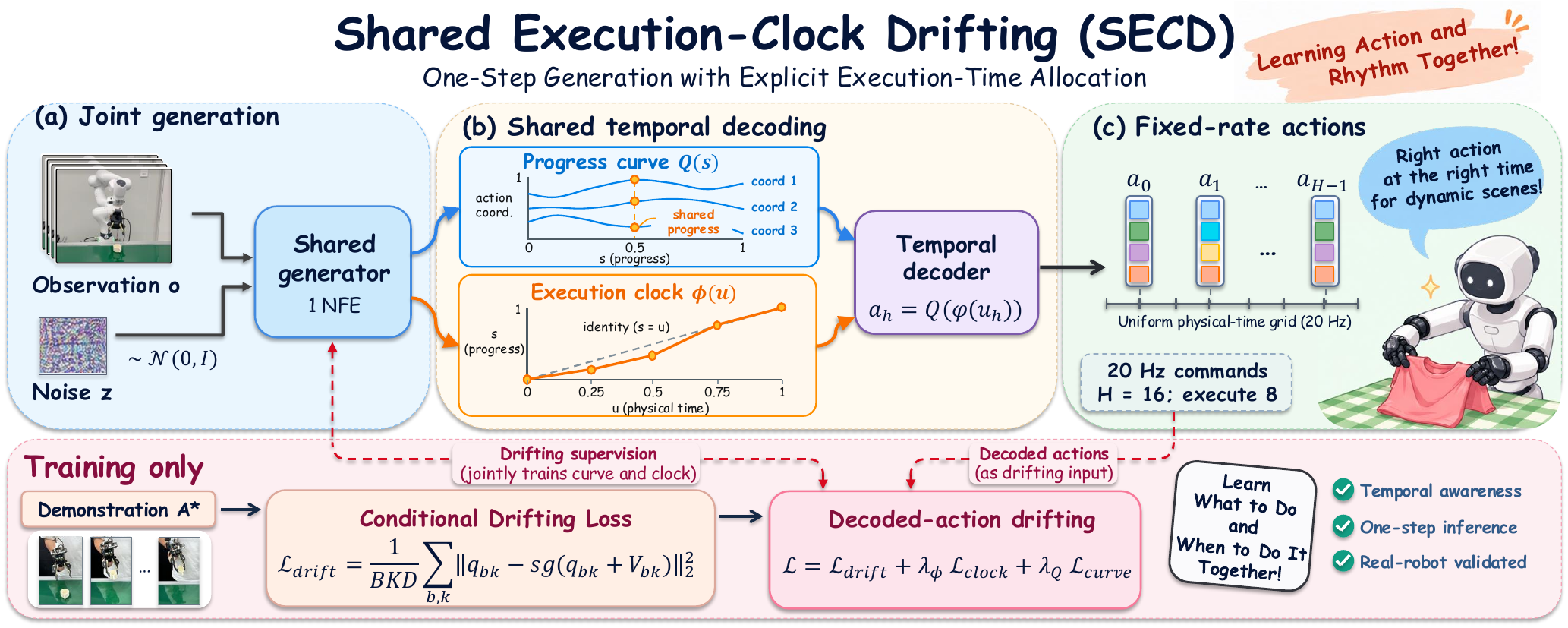}
\caption{\textbf{Shared Execution-Clock Drifting.} A joint generator predicts a progress curve and a monotone clock, which decode to fixed-rate actions in one network evaluation. Demonstration alignment and decoded-action drifting jointly train both outputs. Curves and clock profiles are schematic.}
\label{fig:overview}
\end{figure*}

\section{Introduction}
\label{sec:intro}

Conveyor grasping requires reaching a cup before interception becomes impossible, while closely spaced arrivals leave little time between retrievals. Restoring a crumpled shirt must likewise leave time for the remaining folds. Successful manipulation therefore depends on generated actions and the time assigned to their progression.

Our real-robot observations motivate studying these requirements together. Stationary tasks expose inaccurate placements and missed sleeve grasps. Conveyor retrieval adds missed interception opportunities, while crumpled-shirt recovery can exhaust the $90$\,s budget through ineffective adjustments. Differences among one-step conveyor policies further motivate learning execution rhythm alongside action quality, without isolating timing as every failure's cause.

Action-chunk policies jointly predict time-indexed commands~\cite{chi2023diffusion,act}, learning progression and timing implicitly in the same outputs. They can express adaptive behavior, but expose no separate timing variable. Distillation and one-step generative methods reduce generation cost~\cite{consistency,onediffusion,meanflow,mp1}, and drifting supports whole-chunk generation~\cite{drifting,dbp}. We ask: \emph{can explicitly learning how execution time maps to action progress improve a one-step manipulation policy?}

We propose \textbf{Shared Execution-Clock Drifting (\method)}, which explicitly learns execution-time allocation by jointly predicting a progress-indexed action curve and a monotone clock. Both depend on the observation and latent sample. The clock assigns fixed command times to curve locations, where a shared decoder reads the complete action vector. Controller frequency and prediction horizon remain fixed.

Different curve--clock pairs can decode to the same actions. Demonstration-derived alignment addresses this ambiguity by defining progress from normalized action changes and recording its time correspondence. These targets anchor the decomposition without event annotations. Drifting trains the decoded action distribution through both outputs, coupling progression and temporal allocation.

We evaluate four real-robot tasks with inference on NVIDIA Thor, including $90$\,s crumpled-shirt folding and conveyor retrieval. A fixed-clock conveyor comparison tests the complete representation, while Transport and Square ablations examine its interaction with alignment training.

Our contributions are:
\begin{enumerate}
    \item \textbf{Joint learning of action progression and temporal allocation.} We introduce a one-step generator whose explicit execution clock is anchored by demonstration alignment and trained jointly with the progress curve through decoded-action drifting. The architecture and alignment objective together define the proposed temporal representation.
    \item \textbf{Evaluation under real execution constraints.} Across four tasks with inference on NVIDIA Thor, \method attains 77.00\% task-averaged success. The conveyor comparison improves from 79\% to 91\% over a fixed clock, while simulation controls support the joint representation and alignment design.
\end{enumerate}

\section{Related Work}

\subsection{Generative Visuomotor Policies}

Diffusion Policy~\cite{chi2023diffusion} models conditional action sequences for receding-horizon execution. Rectified flow~\cite{liu2022rectified} provides a transport-based formulation, while consistency policies~\cite{consistency}, one-step diffusion distillation~\cite{onediffusion}, shortcut models~\cite{shortcut}, and mean-flow policies~\cite{meanflow,mp1} reduce sampling cost. HybridFlow~\cite{hybridflow} combines a global MeanFlow proposal with re-noising and local refinement in two network evaluations. These methods motivate efficient action generation. \method investigates a progress-based output representation with an explicitly generated execution rhythm.

\subsection{Drifting for Action Generation}

Drifting trains a generator through attraction--repulsion targets and supports feature-space training~\cite{drifting}. DBPO~\cite{dbp} combines a one-step drifting policy with online policy optimization. DriftingVLA~\cite{driftingvla} introduces per-dimension temporal drifting and retains joint generation through a shared network. IDP~\cite{idp} addresses sparse conditional expert samples through observation-related expert geometry. Our proposed modification concerns the progress curve, monotone clock, and temporal decoder. The drifting estimator, whole-chunk interactions, and one-step sampling remain inherited components.

\subsection{Structured Actions and Temporal Warping}

Movement primitives~\cite{dmp} and waypoint extraction~\cite{awe} establish precedents for phase-based and compact motion representations. Spline Policy~\cite{splinepolicy} predicts continuously decodable spline parameters and also constructs a state-dependent execution field. TimewarpVAE~\cite{timewarpvae} jointly learns spatial representations and monotone time warps from full trajectories, with regularization addressing timing degeneracy. Thus, curve--clock decomposition and differentiable temporal decoding have precedents. \method predicts both components from the current observation and latent sample at each policy query, using demonstration alignment and decoded-action drifting to train their joint distribution.

Other methods address different temporal decisions. PACE~\cite{pace} selects how many predicted actions to execute before replanning. SPECTRA~\cite{spectra} regulates phase speed under joint-motion constraints while retaining spectral coefficients. ISR~\cite{isr} resamples demonstrations offline. \method learns an intra-chunk time-to-progress map while keeping command frequency, prediction horizon, and executed-prefix length fixed. Its contribution concerns the generator parameterization and alignment objective.

\section{Shared-Clock Action Generation}
\label{sec:method}

We make execution-time allocation an explicit learning variable through three connected design choices (Fig.~\ref{fig:overview}). Temporal allocation should be observation-conditioned, which motivates a generated clock. Progress should retain its order while its allocation changes, which motivates monotone decoding. The curve and clock need consistent training meanings, which motivates demonstration alignment. Drifting then supplies a joint training signal through the decoded actions. Implementation settings appear in Sec.~\ref{sec:experiments}.

\subsection{A Joint Curve and Clock Generator}

Let $o$ denote the observation history, which contains state inputs in simulation and image-based observations with proprioception on the robot. The policy predicts $H$ future action commands separated by $\Delta t$. Normalized execution time is $u_h=h/(H-1)$ for $h=0,\ldots,H-1$, with a physical horizon $T=(H-1)\Delta t$.

A progress variable $s\in[0,1]$ indexes the generated action curve. It is local to a predicted chunk and is not an estimate of whole-task completion. An observation encoder and joint action head produce curve coefficients and clock logits:
\begin{equation}
 c=E_\theta(o),\qquad
 (C,v)=F_\theta(c,z),\quad z\sim\mathcal N(0,I).
 \label{eq:structured_output}
\end{equation}
Here $C\in\mathbb R^{L\times d_a}$ specifies a curve through a fixed piecewise-linear basis $b(s)\in\mathbb R^L$, which is differentiable away from its knots:
\begin{equation}
 Q(s)=b(s)^{\mathsf T}C\in\mathbb R^{d_a}.
 \label{eq:progress_curve}
\end{equation}
The curve and decoded commands use normalized action coordinates, which are unnormalized before execution. Clock logits $v\in\mathbb R^{H-1}$ allocate progress between command times. Both outputs depend on the same observation and latent sample within one policy call.

\subsection{A Monotone Execution Clock}

The clock maps normalized execution time to progress. We define positive increments with a lower bound $\epsilon$, where $0<\epsilon<1/(H-1)$:
\begin{equation}
 w_h=\epsilon+\bigl(1-(H-1)\epsilon\bigr)
 \frac{\exp(v_h)}{\sum_{j=0}^{H-2}\exp(v_j)}.
 \label{eq:clock_increments}
\end{equation}
With $\phi_0=0$, the clock nodes are
\begin{equation}
 \phi_h=\sum_{j=0}^{h-1}w_j,\quad h=1,\ldots,H-1.
 \label{eq:clock_nodes}
\end{equation}
Thus $\phi_{H-1}=1$ and $\phi_{h+1}>\phi_h$. Piecewise-linear interpolation between $(u_h,\phi_h)$ defines $\phi(u)$ throughout the horizon. Monotonicity preserves the order of points along $Q$; it does not impose a task-specific order of operations.

The decoder evaluates
\begin{equation}
 \hat a_h=Q(\phi_h),\qquad h=0,\ldots,H-1.
 \label{eq:clock_decode}
\end{equation}
The scalar $\phi_h$ is shared across action coordinates. Larger increments traverse more progress between commands, while smaller increments allocate more time to a curve portion. Controller frequency and prediction horizon remain fixed. In bimanual tasks, different coordinate functions can move or remain constant at the same progress, so sharing does not require equal velocities or simultaneous actions. The structural change is explicit progress-to-time factorization.

\subsection{Anchoring the Decomposition with Demonstrations}

Without a convention, the factorization is not identifiable. For any strictly increasing bijection $\psi:[0,1]\rightarrow[0,1]$, the two pairs
\begin{equation}
 (Q,\phi)\quad\text{and}\quad
 (Q\circ\psi^{-1},\,\psi\circ\phi)
 \label{eq:reparameterization}
\end{equation}
produce the same continuous action sequence. A learned clock therefore needs an anchor if its values are to have a consistent interpretation across demonstrations.

We propose a training-only progress convention based on accumulated changes in normalized action commands. Given a demonstrated chunk $A^*=[a_0^*,\ldots,a_{H-1}^*]$ and fixed dataset normalizer $\mathcal N$, define
\begin{equation}
 r_h=\left\|W\bigl(\mathcal N(a_{h+1}^*)-
 \mathcal N(a_h^*)\bigr)\right\|_2,
 \label{eq:progress_distance}
\end{equation}
where $W=I$ after dataset normalization in our experiments. This convention measures normalized command changes, not physical speed, energy, or semantic task progress.

For a positive floor $\eta$, the target clock increments are
\begin{equation}
 w_h^*=\epsilon+\bigl(1-(H-1)\epsilon\bigr)
 \frac{r_h+\eta}{\sum_{j=0}^{H-2}(r_j+\eta)}.
 \label{eq:target_clock}
\end{equation}
Cumulative sums define $\phi_h^*$. Interpolating $(\phi_h^*,\mathcal N(a_h^*))$ and sampling at fixed progress locations gives the curve target. The positive floor makes progress well-defined when action changes vanish. This convention anchors targets without proving a unique learned factorization. Nearly constant curve portions represent holds. Temporal allocation alone cannot repair an unsuitable action sequence.

\begin{figure*}[t]
\centering
\includegraphics[width=\textwidth]{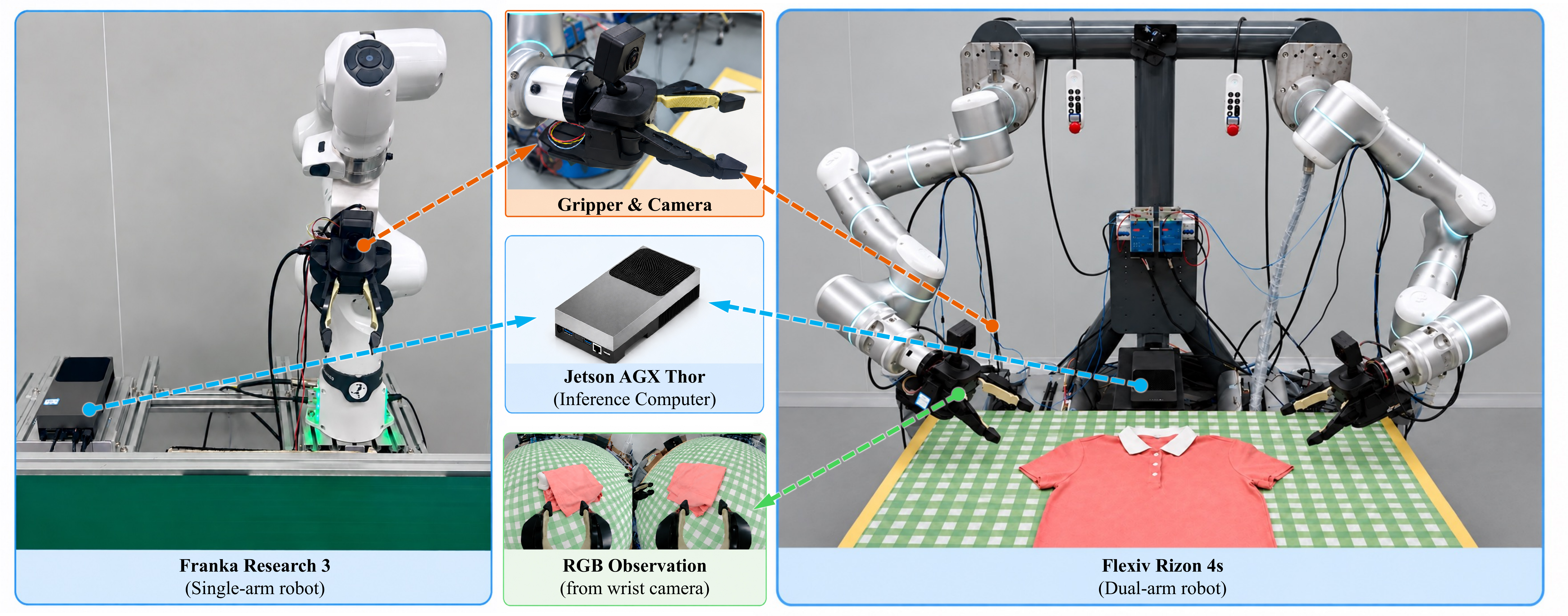}
\caption{\textbf{Real-robot hardware setup.} A Franka Research 3 arm performs conveyor retrieval (left). The Flexiv Rizon 4s setup (right) uses one arm for cup placement and both arms for shirt manipulation. The insets show the gripper and wrist camera, example RGB observations, and the NVIDIA Jetson AGX Thor computer on which all compared policies run inference.}
\label{fig:hardware_setup}
\end{figure*}

\subsection{Training through Temporal Decoding}

Let $\hat A_{bk}$ be the decoded action sequence for observation $b$ and latent sample $k$. We apply the established conditional drifting estimator~\cite{drifting,dbp} to these normalized decoded sequences. Writing $q_{bk}$ for its scaled action coordinates and $V_{bk}$ for its detached field, the inherited frozen-target loss is
\begin{equation}
 \mathcal L_{\mathrm{drift}}=
 \frac{1}{BKD}\sum_{b,k}
 \left\|q_{bk}-\operatorname{sg}(q_{bk}+V_{bk})\right\|_2^2,
 \label{eq:inherited_drift}
\end{equation}
where $D=Hd_a$. Candidate sampling, attraction--repulsion interactions, and normalization follow the cited estimator. They are not introduced as new components of \method.

The proposed auxiliary terms anchor the clock and curve:
\begin{align}
 \mathcal L_{\mathrm{clock}}&=\frac{1}{H-2}
 \sum_{h=1}^{H-2}(\phi_h-\phi_h^*)^2,
 \label{eq:clock_loss}\\
 \mathcal L_{\mathrm{curve}}&=\frac{1}{Jd_a}
 \sum_{j=1}^{J}\|Q(s_j)-Q^*(s_j)\|_2^2,
 \label{eq:curve_loss}
\end{align}
where $H>2$ and the $s_j$ are fixed progress samples. For $H=2$, the clock is fixed by its endpoints and the clock loss is omitted. These terms are averaged across the minibatch and generated samples. The complete training objective is
\begin{equation}
 \mathcal L=\mathcal L_{\mathrm{drift}}+
 \lambda_\phi\mathcal L_{\mathrm{clock}}+
 \lambda_Q\mathcal L_{\mathrm{curve}}.
 \label{eq:training_objective}
\end{equation}
Pointwise curve and clock supervision can reduce diversity across valid alternatives. The weights and their controls must therefore be reported rather than assuming that the factorization preserves multimodality without qualification.

\subsection{Principle: Drifting through a Shared Execution Clock}
\label{sec:clock_principle}

The DBP backbone of DBPO~\cite{dbp} directly parameterizes time-indexed action output. Our composite map $\hat a_h=Q_C(\phi_v(u_h))$ exposes progression and temporal allocation as intermediate variables. Both policies return $H\times d_a$ commands. At $L=H$, a uniform clock gives $\hat A=C$, so the representation includes arbitrary discrete command sequences. The proposed change is a parameterization and training bias, without a claim of a larger action space.

\textbf{Curve and timing variations.} Suppress the observation and sample indices and consider a differentiable point of the decoder. Its first-order action variation is
\begin{equation}
 \delta\hat a_h=b(\phi_h)^{\mathsf T}\delta C
 +Q'_C(\phi_h)\,\delta\phi_h
 +o(\|\delta C\|+\|\delta\phi\|).
 \label{eq:action_variation}
\end{equation}
For fixed $C$, a timing variation moves the action vector along the local tangent $Q'_C(\phi_h)$. Curve-coefficient variations can have both tangential and normal components, so Eq.~\eqref{eq:action_variation} is not an orthogonal decomposition. The overlap motivates the target convention in Eq.~\eqref{eq:target_clock}.

\textbf{The drift-induced clock signal.} Let $\sigma>0$ denote the detached scale used by the inherited estimator, so that $q=\operatorname{vec}(\hat A)/\sigma$. Write $V_h$ for the corresponding $d_a$-dimensional block of the frozen field and $\kappa=2/(BKD\sigma)$. For this sample's contribution to Eq.~\eqref{eq:inherited_drift}, the partial derivatives are
\begin{align}
 \nabla_C\mathcal L_{\mathrm{drift}}
 &=-\kappa\sum_h b(\phi_h)V_h^{\mathsf T},
 \label{eq:curve_drift_gradient}\\
 \frac{\partial\mathcal L_{\mathrm{drift}}}{\partial v_j}
 &=-\kappa\sum_h
 \underbrace{\langle Q'_C(\phi_h),V_h\rangle}_{\text{timing signal at }h}
 \frac{\partial\phi_h}{\partial v_j}.
 \label{eq:clock_drift_gradient}
\end{align}
These expressions follow by differentiating the prediction while holding the drift target and scale fixed. The inner product aggregates all action dimensions into a scalar timing signal. A field component orthogonal to $Q'_C(\phi_h)$ contributes no clock signal at that time, although it can still supervise the curve. On a constant segment, clock learning from decoded actions vanishes locally and clock supervision provides an additional signal.

Let $\pi_j=\operatorname{softmax}(v)_j$ and $\alpha=1-(H-1)\epsilon$. The clock Jacobian is explicit:
\begin{equation}
 \frac{\partial\phi_h}{\partial v_j}
 =\alpha\pi_j\left(\mathbf 1_{\{j<h\}}
 -\sum_{i=0}^{h-1}\pi_i\right).
 \label{eq:clock_jacobian}
\end{equation}
Changing one logit redistributes progress through the normalized increments while retaining the clock endpoints. Thus Eq.~\eqref{eq:clock_drift_gradient} couples the action-space signal to a horizon-wide temporal allocation. It does not prescribe independent per-dimension timing.

These output derivatives describe how the inherited field trains the composite generator. Shared network parameters and auxiliary losses combine curve and clock signals, so parameter updates need not produce isolated time shifts or exact steps along $V$. The identities imply no convergence or task-success guarantee.

\subsection{Implementation and One-Pass Deployment}

Shared encoder and U-Net features feed jointly trained linear curve and clock heads. The curve head produces $L$ progress knots, and clock outputs are linearly resampled to $H-1$ logits. Zero initialization gives a uniform initial clock. Piecewise-linear decoding retains gradients through interpolation weights and adds no network evaluation. Mean policy inference time on NVIDIA Thor is 25\,ms.

Decoding precedes unnormalization and robot-interface conversion. The 6D orientation representation requires valid-rotation conversion and degenerate-encoding handling. Simulation retains native incremental commands where applicable, so decoded curves need not correspond to preserved Cartesian paths after controller integration.

At deployment, one latent sample generates both outputs without expert-derived targets. The controller executes a short prefix and replans. Clock values reset per prediction and do not represent persistent task-wide phase.

\section{Experiments}
\label{sec:experiments}

The evaluation asks: \textbf{Q1:} How does \method compare with direct-output policies? \textbf{Q2:} How effective is it under real execution constraints on NVIDIA Thor? \textbf{Q3:} How do the temporal representation and alignment training contribute? Four real-robot tasks address Q1 and Q2. Conveyor, Transport, and Square ablations address Q3, alongside four-task simulation benchmarks.

\begin{figure}[!t]
\centering
\includegraphics[width=\columnwidth]{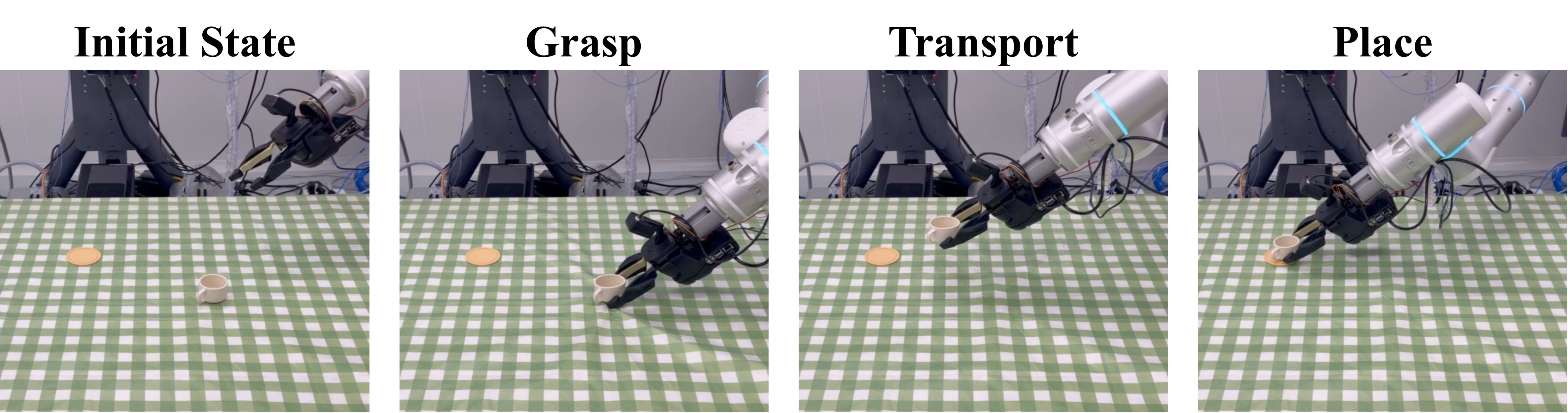}
\caption{\textbf{Cup Place: placing a cup on a coaster.} A single arm grasps and transports a stationary cup to a coaster whose position varies across trials.}
\label{fig:cup_place}
\end{figure}

\begin{figure*}[t]
\centering
\includegraphics[width=\textwidth]{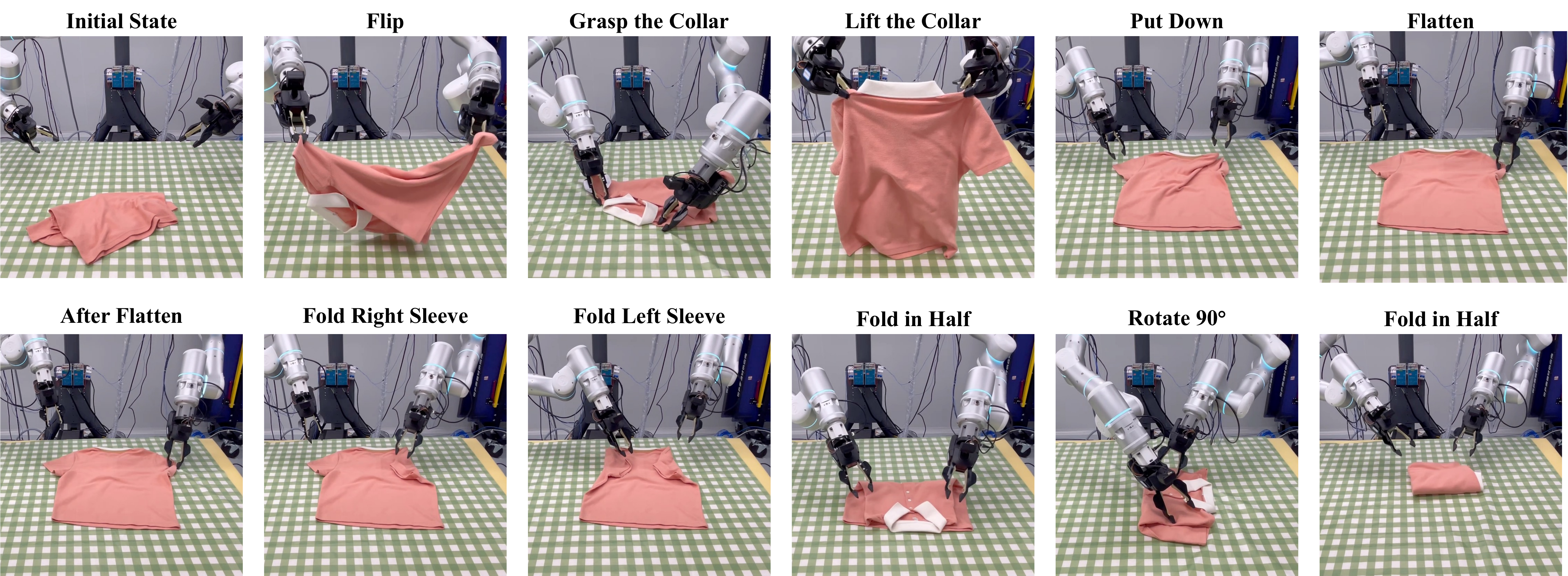}
\caption{\textbf{Shirt folding from flat and crumpled states.} The upper row shows recovery from a crumpled state through flipping, collar grasping and lifting, lowering, and flattening. The lower row shows right- and left-sleeve folding, a half-fold, a $90^\circ$ rotation, and a second half-fold. Fold (Crumpled) requires the full sequence within $90$\,s; Fold (Flat) starts from the flattened state.}
\label{fig:shirt_folding}
\end{figure*}

\subsection{Real-Robot Tasks and Data}

\textbf{Hardware.} Figure~\ref{fig:hardware_setup} shows the Franka Research 3 conveyor setup and the Flexiv Rizon 4s setup, which uses one arm for cup placement and both arms for shirt manipulation. Wrist cameras provide RGB observations. All compared real-robot policies run inference on NVIDIA Jetson AGX Thor with common robot-side deployment settings within each task.

\begin{figure*}[t]
\centering
\includegraphics[width=\textwidth]{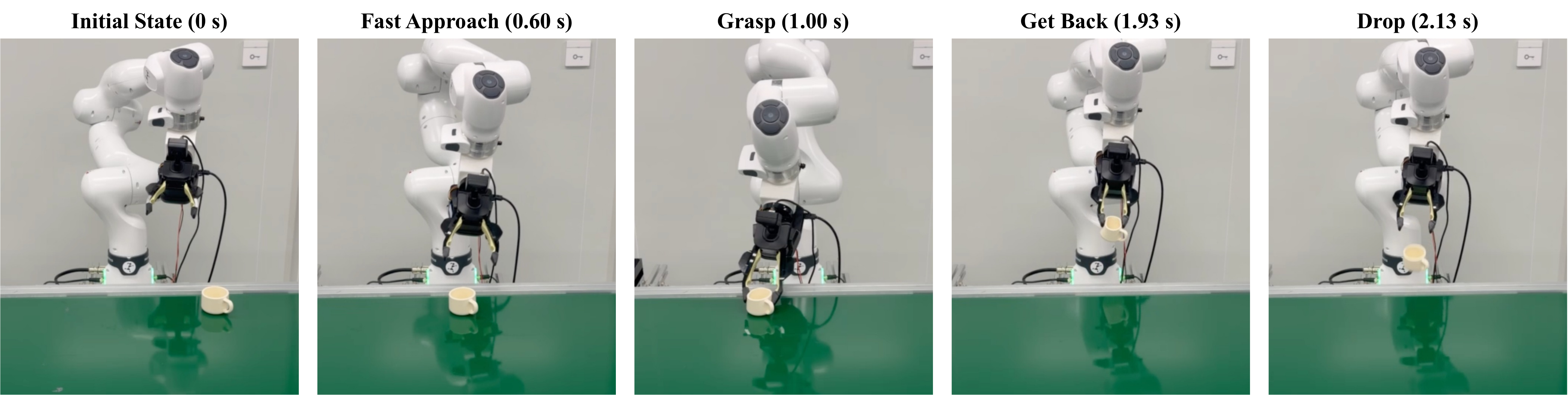}
\caption{\textbf{Conveyor Cup: retrieving a moving cup.} The robot approaches and grasps a cup on a conveyor moving at $16$\,m/min before withdrawing and releasing it. The timestamps indicate elapsed time in the illustrated execution sequence.}
\label{fig:conveyor_cup}
\end{figure*}

\textbf{Tasks.} The four tasks expose different operational requirements:
\begin{enumerate}
    \item \textbf{Cup Place (Fig.~\ref{fig:cup_place}).} A single arm grasps a stationary cup and places it on a randomly positioned coaster. Success requires aligning the cup and coaster centers. Grasp acquisition and release placement are analyzed separately.
    \item \textbf{Fold (Flat) (Fig.~\ref{fig:shirt_folding}, lower row).} Two arms fold the right and left sleeves of a flat shirt, fold it in half, rotate it by $90^\circ$, and fold it in half again. Success is judged by the final configuration after both half-folds. Sleeve grasping tests local operation quality within a longer sequence.
    \item \textbf{Fold (Crumpled) (Fig.~\ref{fig:shirt_folding}, both rows).} A casually tossed shirt starts in a random crumpled configuration. The robot must first flatten it and then complete the same folding sequence within $90$\,s. Exceeding the deadline is a failure. This task also tests recovery from unsuitable grasps, which a shared clock alone does not guarantee.
    \item \textbf{Conveyor Cup (Fig.~\ref{fig:conveyor_cup}).} A single arm grasps a cup moving on a $16$\,m/min conveyor and deposits it into a basket. The approach and grasp must fit a very short interception window. The task tests whether the generated progression and its execution rhythm fit target motion.
\end{enumerate}

\textbf{Demonstrations.} Within each task, compared policies use the same dataset and observation-encoder architecture. The established datasets contain $2$\,h for Cup Place, $10$\,h each for Fold (Flat) and Fold (Crumpled), and $3$\,h for Conveyor Cup. The proposed progress targets are derived from these existing action sequences, using the same normalizer as policy training. No additional event or task-phase annotation is assumed.

\subsection{Comparisons and Implementation Protocol}

\textbf{Policy comparisons.} DP~\cite{chi2023diffusion} uses 16 inference evaluations. OneDP~\cite{onediffusion} distills a teacher configured for 16-step DDIM sampling into a one-evaluation student. Native Drifting~\cite{drifting} denotes our local direct-output policy. Its supplied implementation generates $K=8$ candidates per observation, flattens each full horizon to $Hd_a$ coordinates, and applies drifting with the paired expert sequence as the positive reference and generated candidates as repulsive references. It uses the same drifting-loss routine as \method, with one-sample, one-evaluation deployment. \method adds the curve--clock parameterization and alignment losses. All methods share task datasets and encoder architecture.

\textbf{Fixed-clock ablation.} We additionally train and evaluate a conveyor variant with $\phi_h=h/(H-1)$. It retains the backbone and curve head but removes the learned clock head and clock-alignment loss, and uses uniform-progress curve targets. At $L=H$, uniform decoding samples the curve coefficients directly. This control tests the full aligned temporal representation rather than the clock branch alone.

\textbf{Configuration.} The real-robot configuration uses two observation steps, $224\times224$ RGB inputs, a DINOv3 ViT-B/16 encoder~\cite{dinov3}, and a conditional 1D U-Net with widths $[256,512,1024]$. The horizon is $H=16$, with 10 action coordinates per arm: relative position, 6D orientation, and a scalar gripper command.

\textbf{Training configuration.} The decoder uses piecewise-linear interpolation with $L=J=H$, $\epsilon=\eta=10^{-4}$, $W=I$ in normalized coordinates, and $\lambda_\phi=\lambda_Q=1$. Drifting uses $K=8$ and temperatures $\{0.02,0.05,0.2\}$. The supplied configurations specify 200 epochs, batch size 192, seed 42, and 5\% validation data. AdamW uses generator/encoder learning rates $10^{-4}/10^{-5}$, $\beta=(0.95,0.999)$, weight decay $10^{-6}$, cosine decay, and 2,000 warmup steps. EMA uses power 0.75 and maximum decay 0.9999. OneDP freezes its teacher-initialized encoder; the drifting policies train it jointly.

\textbf{Temporal alignment and deployment.} WebDataset training uniformly samples a 0--4-step offset, limited by available sample length. Images stay fixed while proprioception and actions shift together. Actions begin at the last proprioceptive timestep, and clock targets use the shifted action window. Relative poses retain the latest-image reference. All policies run on NVIDIA Thor at a command rate of 20\,Hz ($\Delta t=50$\,ms), executing eight steps between replans (nominally 0.4\,s) under common robot-side settings. Code and training configurations will be released.

\begin{table*}[t]
\caption{Real-robot success (\%) and mean policy inference time on NVIDIA Thor. Each policy is evaluated on 100 Cup Place, 50 Fold (Flat), 50 Fold (Crumpled), and 100 Conveyor Cup trials. Avg. is the unweighted mean across tasks. Fold (Crumpled) requires flattening and full folding within $90$\,s; Conveyor Cup uses a belt speed of $16$\,m/min. Bold marks the highest success in each column.}
\label{tab:real}
\centering
\begin{tabular}{lccccccc}
\toprule
Method & NFE & Lat. (ms) & Cup Place & Fold (Flat) & Fold (Crumpled) & Conveyor Cup & Avg. \\
\midrule
DP~\cite{chi2023diffusion} & 16 & 252 & \textbf{98.0} & \textbf{74.0} & 30.0 & 0.0 & 50.50 \\
OneDP~\cite{onediffusion} & 1 & 26 & 71.0 & 38.0 & 22.0 & 31.0 & 40.50 \\
Native Drifting~\cite{drifting} & 1 & 24 & 83.0 & 48.0 & 26.0 & 56.0 & 53.25 \\
\method (ours) & 1 & 25 & 95.0 & 68.0 & \textbf{54.0} & \textbf{91.0} & \textbf{77.00} \\
\bottomrule
\end{tabular}
\end{table*}

\subsection{Real-Robot Results and Operation Analysis}

\textbf{Overall performance.} Table~\ref{tab:real} reports 77.00\% task-averaged success for \method. Gains over Native Drifting are 12, 20, 28, and 35 percentage points in task order. \method leads the evaluated one-step policies on every task and all policies on the two tasks with explicit timing constraints.

\textbf{Statistical uncertainty.} The two-sided 95\% Wilson intervals for \method are $[88.8,97.8]\%$, $[54.2,79.2]\%$, $[40.4,67.0]\%$, and $[83.8,95.2]\%$ in table order. These quantify trial-level uncertainty, not training-seed variability. The following analysis discusses task-level failure modes illustrated in the figures.

\begin{figure}[!t]
\centering
\includegraphics[width=\columnwidth]{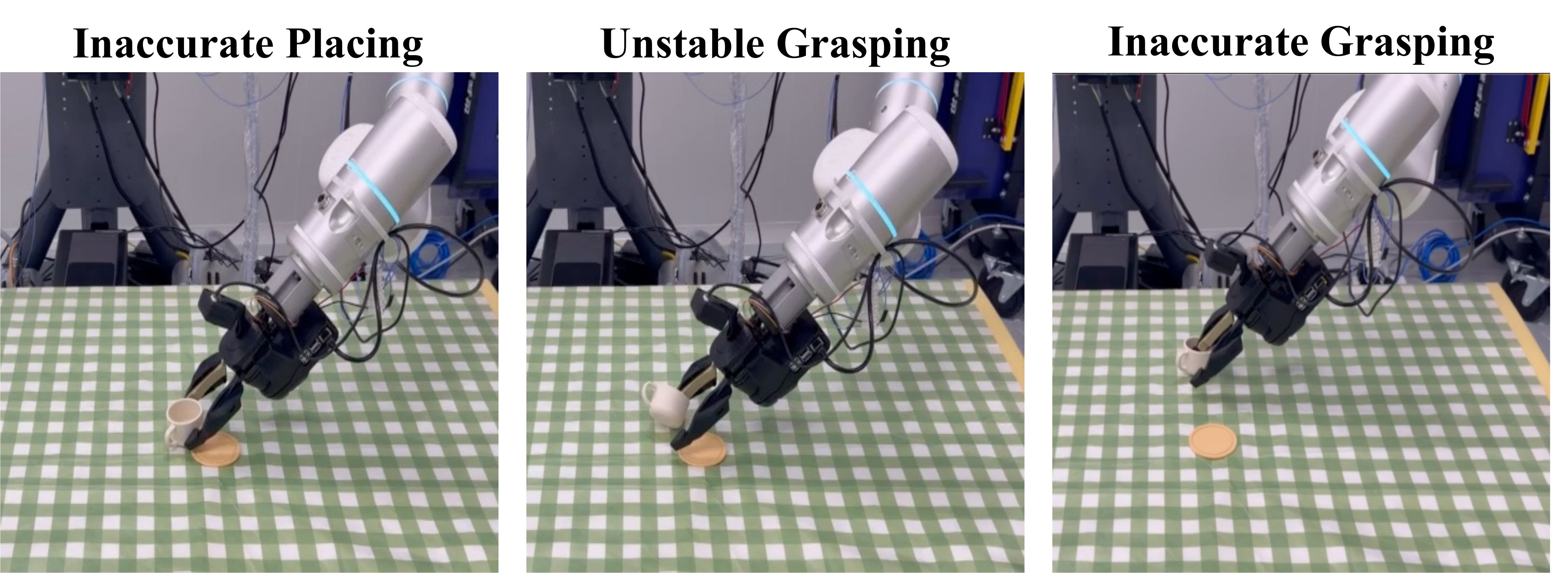}
\caption{\textbf{Cup grasping and placement failures.} The examples show inaccurate placement (left), unstable grasping (center), and inaccurate grasping (right).}
\label{fig:cup_failures}
\end{figure}

\textbf{Cup acquisition and placement.} \method reaches 95\% success, close to DP's 98\% and above Native Drifting's 83\% and OneDP's 71\%. This basic task mainly requires reliable acquisition and accurate release. Baseline observations identified empty initial grasps with OneDP and releases too far from the coaster (Fig.~\ref{fig:cup_failures}). The shared-clock policy retains high precision, although the aggregate counts do not identify the causes of its five failures.

\begin{figure}[!t]
\centering
\includegraphics[width=\columnwidth]{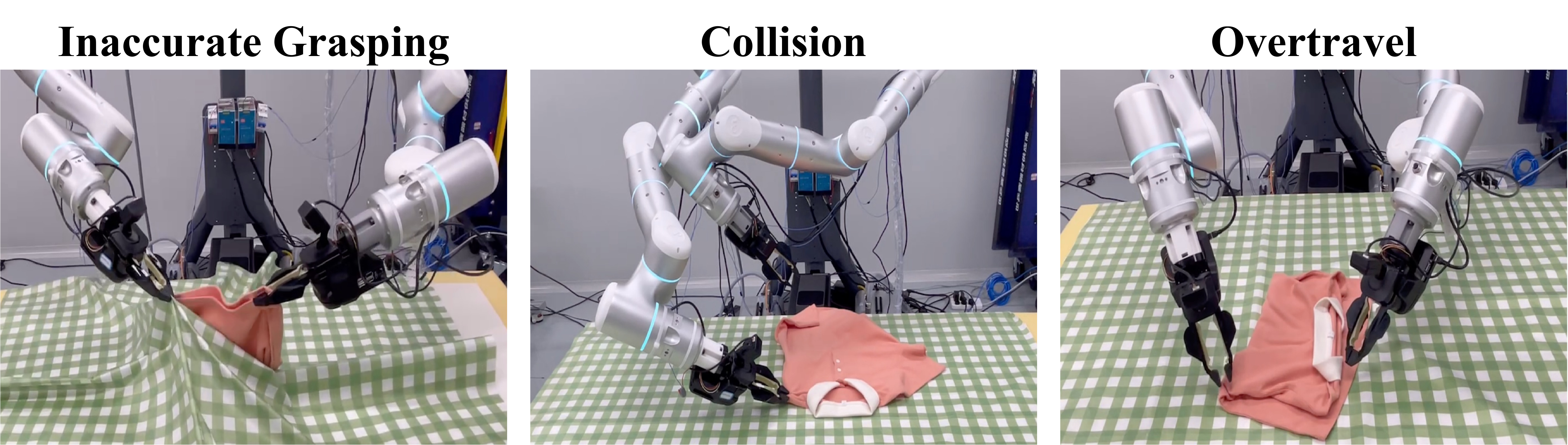}
\caption{\textbf{Failures during shirt manipulation.} The examples illustrate inaccurate grasping, collision, and overtravel that disrupt the manipulation sequence.}
\label{fig:shirt_failures}
\end{figure}

\textbf{Sleeve grasping and folding.} \method completes 34/50 trials (68\%), compared with 74\% for DP, 48\% for Native Drifting, and 38\% for OneDP. This task emphasizes action quality because a missed sleeve grasp can prevent subsequent folds (Fig.~\ref{fig:shirt_failures}). The 20-point gain over Native Drifting extends the advantage to multi-stage manipulation without a deadline, while DP's remaining lead indicates room for improving fine manipulation.

\textbf{Recovery within $90$\,s.} \method completes 27/50 trials (54\%), exceeding DP's 30\%, Native Drifting's 26\%, and OneDP's 22\%. The ranking changes from flat-shirt folding, where DP leads. Crumpled starts add variable recovery work and require flattening early enough to finish both folds. Observations showed that grasping multiple fabric layers can cause ineffective adjustment cycles. The result supports the complete policy under this combined demand for action quality and timely completion. Further mechanism analysis is discussed in Sec.~\ref{sec:limitations}.

\begin{figure}[!t]
\centering
\includegraphics[width=\columnwidth]{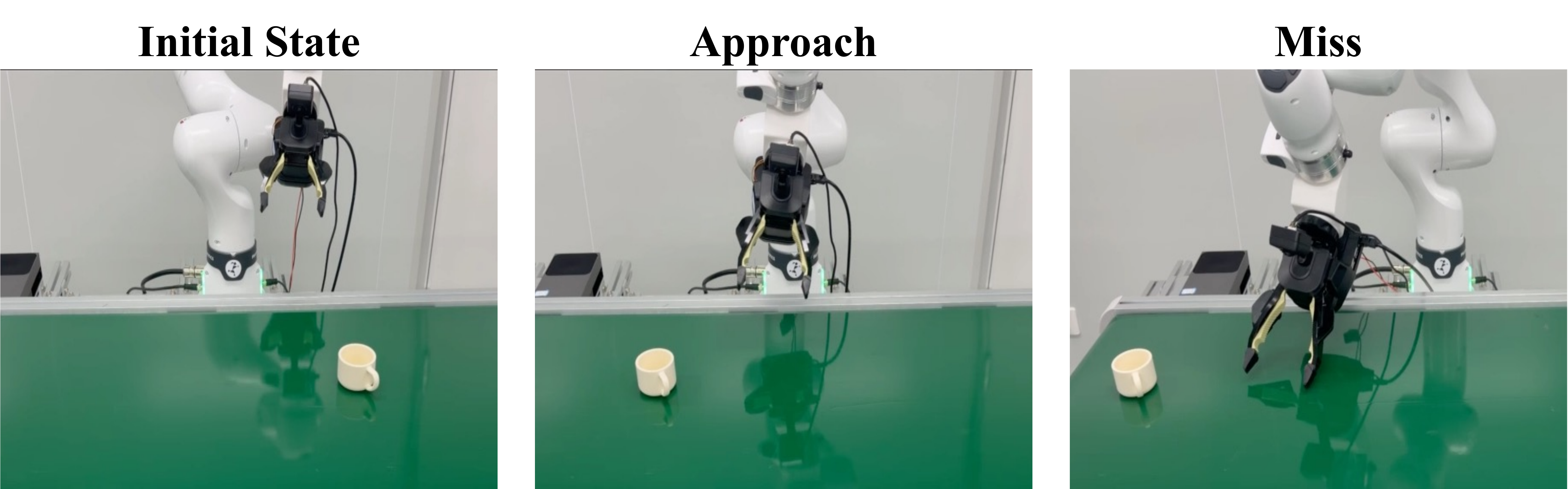}
\caption{\textbf{Missed conveyor grasp with DP (16 NFE).} The cup continues moving during the approach, and the gripper fails to intercept it.}
\label{fig:dp16_conveyor_failure}
\end{figure}

\textbf{Moving-cup interception.} \method succeeds in 91/100 trials on the $16$\,m/min conveyor, compared with 56\% for Native Drifting, 31\% for OneDP, and 0\% for DP. Its 35-point lead over Native Drifting is the largest task-wise gain. The approach and grasp must fit a brief interception opportunity, which the illustrated DP sequence misses (Fig.~\ref{fig:dp16_conveyor_failure}). The strong separation among one-step policies shows that a single network evaluation alone is insufficient. The fixed-clock ablation below further examines the temporal representation.

\begin{table*}[t]
\caption{State-based RoboMimic PH success: best (final-five-checkpoint average); Avg.: four-task mean. $\dagger$: IDP Table 2~\cite{idp}, averaging CNN/Transformer backbones, three seeds, and 50 initial conditions; NFE follows the source. Native Drifting denotes its ``Naive Drifting.'' \method: local U-Net, 100 rollouts/checkpoint.}
\label{tab:robomimic}
\centering
\begin{tabular}{l c ccccc}
\toprule
Method & NFE & Lift & Can & Square & Transport & Avg. \\
\midrule
DP~\cite{chi2023diffusion}$^{\dagger}$ & 100 & 1.00 (1.00) & 1.00 (0.98) & 0.96 (0.92) & 0.73 (0.62) & 0.92 (0.88) \\
Flow~\cite{flowmatching}$^{\dagger}$ & 9 & 1.00 (1.00) & 1.00 (0.98) & 0.95 (0.90) & 0.74 (0.59) & 0.92 (0.87) \\
MIP~\cite{mip}$^{\dagger}$ & 2 & 1.00 (1.00) & 1.00 (0.98) & 0.96 (0.88) & 0.71 (0.58) & 0.92 (0.86) \\
CP~\cite{consistency}$^{\dagger}$ & 1 & 1.00 (1.00) & 0.99 (0.96) & 0.88 (0.82) & 0.32 (0.24) & 0.80 (0.76) \\
MP1~\cite{mp1}$^{\dagger}$ & 1 & 1.00 (1.00) & 1.00 (0.97) & 0.96 (0.88) & 0.52 (0.39) & 0.87 (0.81) \\
Native Drifting~\cite{drifting}$^{\dagger}$ & 1 & 1.00 (1.00) & 0.99 (0.96) & 0.97 (0.90) & 0.68 (0.58) & 0.91 (0.86) \\
IDP~\cite{idp}$^{\dagger}$ & 1 & 1.00 (1.00) & 1.00 (0.97) & 0.95 (0.88) & 0.72 (0.58) & 0.92 (0.86) \\
\method (ours) & 1 & 1.00 (1.00) & 1.00 (0.98) & 0.97 (0.90) & 0.78 (0.68) & 0.94 (0.89) \\
\bottomrule
\end{tabular}
\end{table*}

\subsection{Fixed-Uniform-Clock Conveyor Ablation}
\label{sec:clock_ablation}

Under the same data, training budget, and deployment settings, the separately trained fixed-clock variant retrieves 79/100 cups, compared with 91/100 for complete \method, a 12-percentage-point gain. Both evaluations use the $16$\,m/min conveyor and deliberately include closely spaced cup pairs. Success is counted per cup, not per pair. The two-sided 95\% Wilson intervals are $[70.0,85.8]\%$ and $[83.8,95.2]\%$, respectively.

\textbf{Closely spaced arrivals.} In observed sequences with two cups arriving in close succession, the fixed-clock policy failed to complete retrieval, while \method completed such sequences. This is consistent with the benefit of adapting execution rhythm to successive interception opportunities. It does not establish a minimum arrival gap or a condition-specific success rate. The ablation supports the complete aligned temporal representation. The cup-level intervals do not model dependence between successive cups.

\subsection{Complementary State-Based Simulation}

State-based experiments use the RoboMimic proficient-human Lift, Can, Square, and Transport datasets~\cite{robomimic}, retaining the previous setup. The U-Net has widths $[256,512,1024]$, kernel size 5, eight normalization groups, and generation-index embedding dimension 256. It conditions on two observation steps, predicts 16 actions, and executes eight before replanning. Training uses seed 1 and batch size 100 for 2,000 epochs of 100 updates, AdamW at $10^{-4}$ with weight decay $10^{-6}$, cosine decay with 500 warmup steps, and EMA power 0.75. Each evaluation uses 100 rollouts and runs every 50 epochs, with early termination on task success.

Table~\ref{tab:robomimic} reports best success and final-five-checkpoint averages. The task means for \method are 0.9375 and 0.8900, displayed as 0.94 (0.89). Transport improves numerically from the cited Native Drifting result of 0.68 (0.58) to 0.78 (0.68). Square matches its 0.97 (0.90), while DP reports a final-five average of 0.92. CP denotes Consistency Policy, distinct from real-robot OneDP. Literature comparisons do not isolate architectural effects.

The decoder retains each benchmark's action representation and output conversion. Decoding incremental commands need not preserve the accumulated physical path.

\subsection{Representation and Alignment Ablations}
\label{sec:representation_ablation}

Additional local Transport and Square ablations report results aggregated across training seeds, with 100 rollouts per checkpoint evaluation and otherwise unchanged settings. Table~\ref{tab:representation_ablation} compares direct-output drifting (A), the fixed-clock variant with action MSE (B), \method with both alignment weights set to zero (C), and complete \method (D). C and D share the same architecture and parameter count. At $L=H$, B's uniform decoder and curve target reduce to direct action prediction and action MSE.

\begin{table}[t]
\caption{Transport and Square ablations: success rates aggregated across training seeds. Best and Final denote best-checkpoint success and the final-five-checkpoint average, respectively.}
\label{tab:representation_ablation}
\centering
\setlength{\tabcolsep}{3pt}
\begin{tabular}{lcccc}
\toprule
& \multicolumn{2}{c}{Transport} & \multicolumn{2}{c}{Square} \\
\cmidrule(lr){2-3}\cmidrule(lr){4-5}
Variant & Best & Final & Best & Final \\
\midrule
A: Direct drifting & 0.68 & 0.58 & \textbf{0.97} & 0.90 \\
B: Uniform clock + MSE & 0.74 & 0.60 & 0.96 & \textbf{0.92} \\
C: \method w/o alignment & 0.70 & 0.56 & \textbf{0.97} & 0.88 \\
D: Complete \method & \textbf{0.78} & \textbf{0.68} & \textbf{0.97} & 0.90 \\
\bottomrule
\end{tabular}
\end{table}

On Transport, action MSE improves B over A by 6/2 percentage points (Best/Final), whereas C changes A by $+2/-2$ points. Complete \method exceeds the same-architecture C by 8/12 points and B by 4/8 points. On Square, D matches C's best success and improves its Final by 2 points, while trailing B's Final by 2 points. Thus, alignment benefits the learned-clock architecture on both tasks, while the additional gain over uniform-clock action regression depends on the task. These aggregated results neither isolate the two alignment losses nor quantify cross-seed dispersion.

\section{Limitations and Future Work}
\label{sec:limitations}

The results support the complete policy but do not identify which stages advance faster or slower. Future work will relate predicted clocks to observations and execution events. Transport and Square controls support alignment within the same architecture but do not isolate its two losses. Generalization to other tasks and per-seed variability analysis remain future work.

Progress and time admit equivalent parameterizations. Demonstration alignment anchors targets without establishing a unique representation. Its action-change metric depends on normalization and does not measure semantic progress. Clock non-uniformity alone does not establish useful adaptation, which also depends on the curve. Basis resolution, positive increments, and auxiliary losses can limit fidelity or diversity. Inherited drifting does not resolve missing conditional expert modes.

A monotone clock preserves curve traversal order without guaranteeing dynamic feasibility. Controller interpolation, actuator dynamics, and observation delay affect execution. The local clock provides neither task-level recovery planning nor a deadline guarantee. Table~\ref{tab:robomimic}'s baseline rows are literature results, whereas Table~\ref{tab:representation_ablation} reports local controls. Conveyor evaluation covers one speed, and real-robot trial intervals do not quantify training-seed variability.

\section{Conclusion}

Shared Execution-Clock Drifting explicitly learns temporal allocation through an aligned curve--clock parameterization. It achieves 77.00\% task-averaged real-robot success on NVIDIA Thor, including 54\% in deadline-constrained folding and 91\% in conveyor retrieval. The conveyor and simulation ablations support the combined representation and alignment design. RoboMimic evaluation yields 93.75\% task-averaged best success and 89.00\% final-five-checkpoint average success.

\end{document}